\documentclass[runningheads]{llncs}

\usepackage{graphicx}
\usepackage{paralist}
\usepackage{tabularx}
\usepackage{comment}
\usepackage{enumitem}
\usepackage{enumitem}
\usepackage{comment}
\usepackage{subcaption}
\usepackage{wrapfig}
\usepackage{threeparttable}
\usepackage{makecell}
\usepackage{etex}
\usepackage{hyperref}
\usepackage{diagbox}
\usepackage[utf8]{inputenc}
\usepackage[T1]{fontenc}
\usepackage[ngerman,english]{babel}
\usepackage{times}
\usepackage{multirow}
\usepackage[dvipsnames,table,x11names]{xcolor}
\usepackage{booktabs}
\usepackage{paralist}
\usepackage{amssymb, amsfonts}
\usepackage{amsmath}
\usepackage{amsopn}
\usepackage{url}
\usepackage{multirow}
\usepackage{relsize}
\usepackage{xfrac}
\usepackage{pdflscape}
\usepackage{cite}
\usepackage{tabularx}
\usepackage{mdframed}
\usepackage{marginnote}
\usepackage{algorithm}
\usepackage{algorithmicx}
\usepackage[noend]{algpseudocode}
\usepackage{inconsolata}
\usepackage{nicefrac}
\usepackage{xspace}
\usepackage{tabu}
\usepackage{todonotes}
\usepackage{listings}
\usepackage{array} 
\usepackage{ragged2e}

\hypersetup{
	pdftoolbar=true,        % show Acrobat?s toolbar?
	pdfmenubar=true,        % show Acrobat?s menu?
	pdffitwindow=false,     % window fit to page when opened
	pdfstartview={FitH},    % fits the width of the page to the window
	pdftitle={},    % title
	pdfauthor={anonym},     % author
	pdfsubject={},   % subject of the document
	pdfcreator={},   % creator of the document
	pdfproducer={}, % producer of the document
	pdfkeywords={}, % list of keywords
	pdfnewwindow=true,      % links in new window
	colorlinks=true,       % false: boxed links; true: colored links
	linkcolor=Brown,          % color of internal links (change box color
	citecolor=Brown,        % color of links to bibliography
	filecolor=Brown,      % color of file links
	urlcolor=Brown           % color of external links
}

\newcommand{\mypar}[1]{\noindent\textbf{#1.}}

\usepackage[T1]{fontenc}
\usepackage{graphicx}
\begin{document}
\title{Remaining Time Prediction in Outbound Warehouse Processes: A Case Study (Short Paper)}

\titlerunning{Remaining Time Prediction: A Case Study}

\author{
Erik Penther\inst{1} 
\and
Michael Grohs\inst{1}\orcidID{0000-0003-2658-8992} 
\and
Jana-Rebecca Rehse\inst{1}\orcidID{0000-0001-5707-6944}}
\authorrunning{E. Penther et al.}
\institute{University of Mannheim, Mannheim, Germany \\
\email{\{erik.penther@students., michael.grohs@, rehse@\}uni-mannheim.de}}
\maketitle              % typeset the header of the contribution
\begin{abstract}
Predictive process monitoring is a sub-domain of process mining which aims to forecast the future of ongoing process executions. 
One common prediction target is the remaining time, meaning the time that will elapse until a process execution is completed.
In this paper, we compare four different remaining time prediction approaches in a real-life outbound warehouse process of a logistics company in the aviation business. For this process, the company provided us with a novel and original event log with 169,523 traces, which we can make publicly available. 
Unsurprisingly, we find that deep learning models achieve the highest accuracy, but shallow methods like conventional boosting techniques achieve competitive accuracy and require significantly fewer computational resources.
\end{abstract}

\section{Introduction}
\label{sec:introduction}

Process mining is a family of data analysis techniques that aims to provide insights into business processes within organizations \cite{dumas_fundamentals_2018}.
To this end, process mining techniques analyze recorded process executions, so-called traces, which are captured and stored collectively in event logs \cite{dumas_fundamentals_2018}.
A well-established sub-discipline of process mining is Predictive Process Monitoring (PPM), which focuses on predicting the future progression of ongoing traces \cite{di_francescomarino_predictive_2022}. One of the most common prediction targets is the remaining time until completion of traces \cite{gunnarsson_direct_2023}. Remaining time prediction can help to avoid deadline violations, improve operational efficiency, and provide estimates to customers \cite{elyasi_pgtnet_2024}. In recent years, numerous approaches for remaining time prediction have been proposed, which aim to address the challenges of the task such as capturing long-range dependencies and harnessing process perspectives other than the control-flow \cite{elyasi_pgtnet_2024}. 
Many approaches are based on deep learning models \cite{wuyts_sutran_2024}, but traditional machine learning methods, such as boosting, have also been employed \cite{roider_assessing_2024}. 

Given this wide availability of approaches, companies who want to do remaining time prediction typically have to choose a well-suited approach for their application case, based on the characteristics of the process in question \cite{di_francescomarino_genetic_2018}. Relevant characteristics may include process complexity, the availability of static and dynamic data attributes, among other factors \cite{di_francescomarino_predictive_2022}. Often, multiple candidate approach may appear suitable for a given use case, making the selection of an appropriate technique a non-trivial task \cite{di_francescomarino_genetic_2018}.

In this paper, we illustrate this selection of an appropriate remaining time prediction technique by comparing the performance of different approaches for a real-life outbound warehouse process.
For this process, we obtained an event log from a company, which we can make publicly available. 
To this event log, we apply different state-of-the-art remaining time prediction approaches to determine the most suitable one. In particular, we apply three sophisticated deep learning techniques to the outbound warehouse process and assess how accurately they predict the remaining time of ongoing traces. In addition, we employ a rather simple baseline approach based on XGBoost. Our findings indicate that, unsurprisingly, deep learning approaches are more accurate than our simple baseline but also require substantial training effort. 
However, XGBoost performs competitively and even outperforms one deep learning approach. Consequently, it can be a viable alternative since it requires significantly fewer computational resources. 
%Thus, this paper constitutes a real-life case study that shows how applicable existing PPM approaches are to the outbound warehouse process at hand. 

%The remainder of this paper is structured as follows. After our case is presented in \autoref{sec:process}, \autoref{sec:background} provides background knowledge on remaining time prediction. \autoref{sec:research method} elaborates on the method applied in our case study. The results are presented in \autoref{sec:results}. We discuss our findings in \autoref{sec:discussion} and conclude in \autoref{sec:conclusion}.
% 2nd Problem: rarely applied in practice, one main contributing reason is lack of proper vis and also no idea which purpose CC should fulfill (idealerweise Zitat unseres IS-Papers)

% 3rd Problem (contin.): not enough research on vis in CC, practice intransparent about why they use /visualize CC how - Result is an unclear connection between proposed solutions in practice and conformance checking purposes (tasks) - or even stronger: unclear suitability between vis and task 

% 4th solution: To shine light on this problem, we systematically analyze existing CC-visualizations w.r.t. how well they address CC-tasks. In particular ...

\section{Our Case: Outbound Warehouse Process}\label{sec:process}
The process analyzed in this case study originates from a company that provides logistics services for the aviation industry.
The goal is to ensure the efficient supply of aircraft components, thereby focusing on smaller items such as spare parts.
Since the shipment of these smaller items is standardized, the control-flow is relatively straightforward and linear. However, cycle times can vary due to factors such as item type or weight. 
This can cause difficulties in obtaining accurate forecasts for the delivery, but customers require such forecasts to plan the maintenance and repair of aircraft supplies. 

For this process, we obtained an event log with 169,523 traces that we make publicly available in an anonymized form.\footnote{\url{http://figshare.com/articles/dataset/Warehouse_outbound_event_log/29500898}} 
Based on this event log, the process managers seek for a reliable remaining time prediction approach to provide accurate time predictions for downstream tasks that use parts from the warehouse.
Each trace in the log refers to one order that corresponds to one concrete order item.
The desired control-flow consists of seven subsequent activities, of which one is optional.
24 attributes are recorded for each trace, which can be utilized in remaining time prediction.\footnote{The event log does not contain an event attribute for the executing resource due to privacy and data security reasons.} From these 24 attributes, 20 are categorical (e.g., the items' type) and four are numerical (e.g., the shipment weight).
We note that the log has been anonymized due to data privacy reasons. Concretely, activity names have been altered and trace attributes are enumerated and hashed. Nevertheless, the log contains real-life behavior that can be used independently of the fact that activities and attributes have no semantic meaning anymore.

\section{Remaining Time Prediction}
\label{sec:background}

\mypar{Event logs, traces, and prefixes} An event log is a collection of traces, where each trace corresponds to a process execution. A trace consists of a time-ordered sequence of events, each indicating the execution of a specific activity during the process. 
Traces and events can have numerical or categorical attributes. For example, a trace attribute can be the value of shipped goods whereas the executing resource can be an event attribute.
A running trace is referred to as a prefix, which is a sub-sequence of the trace starting from the initial activity. The prefix length indicates how many events from the trace are included. For example, a prefix of length 3 includes the first three events.

%\mypar{Outbound warehouse process} 

\mypar{Remaining time prediction task} 
Remaining time prediction tries to forecast the time that will elapse until an ongoing trace is completed \cite{di_francescomarino_predictive_2022}. Typically, this is done by training a predictive model on a subset of traces, the so-called training set. The performance of the models is then evaluated on the remaining traces, the so-called test set \cite{di_francescomarino_predictive_2022}. Models can utilize all recorded attributes of the event data such as activities or timestamps \cite{di_francescomarino_genetic_2018}.

\section{Case Study Method}
\label{sec:research method}

This section presents the steps we took to assess the suitability of different remaining time prediction approaches for the outbound warehouse process.

\mypar{Pre-Processing}
In the beginning, we pre-processed the event log to ensure that we work with accurate and reliable data. For that, we removed outliers to account for logging issues due to wrong or outdated master data or manual entries. First, we removed traces with logically impossible durations or traces taking over half a year. 
%These are considered unrealistic since almost all orders are processed within a few days. As this can impact the predictive performance of the model, only traces with a duration within the 95th percentile were kept (around 8 days). 
Second, we removed outliers with a weight of more than the 95th percentile since heavier values would suggest the shipment of a whole aircraft engine which is not part of this process. %Realistically, the weight should be less than 1,500 kg per package, as heavier values would suggest the shipment of a whole aircraft engine which is not part of this process. For the same reason, orders with more than 99th percentile of positions were also removed. 
This reduced the number of traces to 130,835, with 1,043,555 events.

In addition, we selected a time frame that contains only the current process version to account for concept drifts, i.e., data from multiple versions of the same process. %This can negatively impact accuracy since the predictive models are trained with data from process versions that are no longer executed, resulting in over- or underestimation of the remaining time in ongoing process executions. 
In particular, a concept drift occurred in the process behavior in May 2024, where some process variants were discontinued whereas others were executed more often.
%This is, among other things, caused by a change in the underlying system, which recorded more events starting in May. 
Therefore, we used only data from June 2024 onward.
This reduced the number of traces to 41,927 consisting of 330,709 events (7.9 events on average).
The average cycle time of traces was 24 hours, with a maximum of 192 hours, indicating quick cycle times.

\mypar{Feature Selection}
Next, we selected the features to be used for prediction. First, we removed uninformative features, striving to keep only informative features in a low-dimensional feature space. This is desired since  high-dimensional feature spaces negatively impact learning algorithms \cite{maimon_data_2010}. 
%Especially in presence of many categorical attributes, as in our case with 20 of 24 features being categorical, more features tend to introduce unnecessary complexity and collinearity. 
We removed three categorical feature with several thousand realizations that likely possess only low predictive power.\footnote{The features cannot be explicated due to privacy reasons. They refer to the categorical features 16, 18, and 20 in the anonymized event log.} Also, we removed two features with only one value, which, by definition, do not have any predictive power.

After that, we selected a subset of features that has potentially larger predictive power based on the Mutual Information (MI) shared between a feature and the remaining time of prefixes in the training set \cite{vergara_review_2014}. All features were ranked, and only those with MI scores above 1 were retained. This reduced the number of features to eleven.

Finally, we created additional features for which the managers of the process know that they have predictive power. These include attributes such as \textit{time since trace started}, \textit{time since last event}, and \textit{day of the week}. Also, we calculated the number of concurrent traces open at the time of the event as a basic inter-case feature. This may help the algorithm to learn capacity utilization since traces share the same resources.

\mypar{Dataset Partitioning}
% Time frame selection, process drift/evolution
We split the event log into a training and test set using a 70-30 split. 
%To mitigate data leakage, we applied strict temporal splitting \cite{weytjens_creating_2022} which allocates a trace to the training set if it has been completed at a defined separation point in time. Consequently, the test set only contains traces which are started after the separation point. Thus, we ensure that there are no traces partially in both training and test set. 
Also, we used the last 10\% of the traces in the training set as the validation set.

\mypar{Predictive Approach Selection}
For finding the most suitable approach for remaining time prediction, we pre-selected four approaches to be tested, based on the most recent developments in remaining time prediction: \begin{inparaenum}[(1)]
    \item A data-aware LSTM approach \cite{gunnarsson_direct_2023} 
    \item A transformer-based approach (SuTraN) \cite{wuyts_sutran_2024} 
    \item A graph transformer-based approach (PGTNet) \cite{elyasi_pgtnet_2024}
    \item An XGBoost-based approach \cite{roider_assessing_2024}
\end{inparaenum}

The former three can be considered state-of-the-art approaches for remaining time prediction, which represent distinct and promising paradigms: LSTMs as one of the first deep learning cells to process sequential data, SuTraN as a rather novel encoder-decoder transformer, and PGTNet with its innovative graph-based event log representations. All three potentially suit our process at hand.
The last approach XGBoost serves as a less sophisticated baseline to show how deep learning relates to other ML approaches.

\mypar{Experimental Setting}
The experiment was conducted using an Apple M1 Pro with 32 GB shared memory in a Python 3.10 environment. The implementations of the deep learning algorithms of LSTM and SuTraN \cite{wuyts_sutran_2024} as well as PGTNet \cite{elyasi_pgtnet_2024} are based on PyTorch and initially utilize NVIDIA CUDA cores. Since these were not available in our setting, the implementations were altered to use available GPU resources.

\mypar{Metrics}
We apply the commonly used Mean Absolute Error (MAE) as evaluation metric for remaining time prediction as it provides a clear accuracy measure\cite{roider_assessing_2024}. %It measures the average absolute difference between the predicted and actual values:

%In addition, we also considered the time needed for training and inference by the predictive approaches.

% Hyperparamter tuning

\mypar{Hyperparameter Optimization}
We optimized hyperparameters of all approaches with a grid search, depending on the approach. Concretely, we optimized these parameters:
\begin{compactenum}[(1)]
    \item LSTM: hidden layer size, no.  of shared layers, no. of dedicated layers, batch size, batch interval, dropout 
    \item SuTraN: hidden layer size, no.  of prefix encoder layers, no. of decoder layers, no. of heads, batch size, batch interval, dropout
    \item PGTNet: positional encoding dimensions, positional encoding times, no. of layers, no. of heads, dropout
    \item XGBoost: no. of estimators, learning rate, subsample, colsample, maximum depth
\end{compactenum}

We selected the best performing hyperparameters for each approach based on the validation set. The next section presents the results of the approaches.

\section{Results}
\label{sec:results}

In this section, we present the results of our case study. %First, we elaborate on overall performance of the used predictive approaches. Then, we assess the performance over time, i.e., across different prefix lengths. Finally, we investigate the approaches performance in early stages of the process.
Our evaluation pipeline including the original and pre-processed event log can be found online.\footnote{\url{https://github.com/ultrawaffle/ppm_remaining_time}} 

\mypar{Overall performance} 
Table~\ref{tab:results} presents the evaluation results of the four different approaches in form of MAE as well as training and inference time. Among the approaches, SuTraN achieves the lowest MAE with 554 minutes, suggesting that it delivers the most accurate predictions. LSTM follows closely with a MAE of 568 minutes, performing only slightly worse than SuTraN. XGBoost has an MAE of 613 minutes, making it the third most accurate approach. In contrast, PGTNet shows a significantly higher MAE of 1390 minutes, likely due to overfitting on training data.

With respect to required time, XGBoost is the fastest to train, requiring only 2 minutes, and has the shortest inference time of 0.10 ms. This suggests that it is highly efficient for quick retraining and real-time predictions. In comparison, LSTM requires a much longer training time of 1.26 hours, but its inference time remains relatively low at 0.63 ms. SuTraN, despite achieving the best predictive performance, has the longest training time of 4.65 hours and a relatively higher inference time of 3.17 ms, indicating that its accuracy comes at the cost of more computational requirements. PGTNet, on the other hand, takes 0.8 hours to train, which is moderate compared to the other models, but its inference time is high at 95.39 ms, making it the least efficient. %The lower training time of PGTNet, although it is a transformer model, can be explained by the fact that early stopping is applied. 

\begin{table}[htb]
\centering
\vspace{-2.5em}
\caption{Average Results of the Models Performances}
\begin{tabular}{llll}
\toprule
\textbf{Approach} & \textbf{MAE (min)} & \textbf{Training time} & \textbf{Inference time} \\ \midrule
LSTM    & 568  & 1.26 h & 0.63 ms  \\
SuTraN  & \textbf{554}  & 4.65 h & 3.17 ms  \\
PGTNet   & 1390 & 0.8 h  & 95.39 ms \\ XGBoost & 613  & \textbf{2 min}  & \textbf{0.10 ms}  \\ \bottomrule
\end{tabular}
\vspace{-1em}
\label{tab:results}
\end{table}

\mypar{Discussion}
We found that some models demonstrate advantages in specific areas but also unraveled challenges that impact their applicability.

\textbf{LSTM} achieves better accuracy than XGBoost. However, LSTMs require longer training times and demand more computational resources. 
Also, static trace attributes cannot be learned dynamically. 
Additionally, in this specific experiment, LSTMs tended to underestimate the remaining time, which could be a bigger problem than overestimation as customers might already need the shipped parts when promised, whereas overestimated times are likely to cause no shortages at the customers' sites. 

\textbf{SuTraN}, a transformer-based model, leverages attention mechanisms to capture long-term dependencies. Among all tested models, SuTraN achieved the best performance in this use case. However, it comes with significantly more computational demands than the other tested models, leading to the longest training time.

\textbf{PGTNet} has demonstrated strong performance in other studies where the dataset is sufficiently complex \cite{roider_assessing_2024}. However, in our use case, PGTNet struggled with overfitting, even after extensive hyperparameter tuning. This suggests that its architecture may be too complex for the given event log. Thus, PGTNet was ultimately deemed unsuitable.

\textbf{XGBoost} is an efficient model, offering advantages in both training and inference speed. It is the quickest out of the tested approaches. 
However, XGBoost does not inherently model temporal dependencies and we have to use specific encoding strategies such as aggregation or index encoding \cite{di_francescomarino_predictive_2022}, potentially leading to information loss. This might explain the worse accuracy in contrast to LSTM and SuTraN.

%Overall, we find that, unsurprisingly, deep learning approaches achieve the highest accuracy but also demand more computational resources and training times.
%However, we trained the models locally in this case study and still achieved training times of below five hours, which likely can be reduced in industrial setting with more powerful servers. 
%Thus, our recommendation to the management of the process is to choose either SuTraN or XGBoost, dependent on whether longer training time can be accepted. 

\section{Conclusion}\label{sec:conclusion}
In this paper, we evaluated the accuracy of four models in predicting the remaining time in process instances of a warehouse outbound process.
Our work has both practical and scientific implications. For practitioners, our study shows that there are alternative predictive approaches among which should be chosen carefully. Especially if periodic re-training is required, shallow approaches like XGBoost might be suited better. Thus, the model choice has to be adapted to the process at hand.
For researchers, our study shows that there is still room for improvement in predictive approaches. In particular, even the state-of-the-art approaches are not able to reduce the MAE below about 9 hours, a large error relative to an average trace duration of only 27 hours. Further, our findings suggest simple architectures suit shorter processes, while other studies found that larger ones handle complex tasks better, indicating potential benefits of hybrid strategies.

Our study is suspect to some limitations.
First, we acknowledge that there are factors that impact processing times that we were not able to quantify. For instance, executing resources were not recorded in the event log. %More concretely, no information regarding the executing resources are available in our case study. Such factors that cannot be quantified pose challenges in making highly accurate process predictions.
Second, the hyperparameter optimization and model testing can always be further refined. Although an extensive tuning process was conducted, the models are highly sensitive to hyperparameters, and more exhaustive tuning could lead to further improvements. %However, given diminishing returns in our grid search, we believe to cover sufficiently many hyperparameter settings.
Finally, we only applied a subset of all remaining time prediction approaches. %Consequently, other approaches might be better suited to this use case. 
Nevertheless, we believe that the four applied models cover a wide range of recent works and a less sophisticated baseline.

%In summary, our case study allows the company at hand to predict the remaining time of process instances as accurately as possible. We believe that case studies like this one benefit the understanding of the progress in research as it can show how organizations can use novel approaches to gain insights into their processes. Such an alignment of scientific progress with practical use cases, process mining can foster collaboration from which both sides profit.
%For that reason, we want apply different approaches in real-life settings, including outcome or next event prediction \cite{di_francescomarino_predictive_2022}, but also other predictive approaches that typically face data imbalance like deviation prediction \cite{grohs2025proactive}. 

Last, we want to stress the fact that we provide a novel public dataset based on a real-life process in the repository referenced above. Although the log is anonymized due to privacy reasons, the recorded behavior is still a detailed depiction of reality. We highly encourage fellow researchers to use this dataset for their own research and, for example, try to discover and characterize the underlying concept drift in the event log. Given that the availability of public event logs is of utmost important for the validity of experimental evaluations, we believe that the publication of the event log can help researchers in the field of process mining when conducting their experiments.

 \bibliographystyle{splncs04}
\bibliography{references}

\end{document}